\documentclass[acmtog]{acmart} 

\usepackage{booktabs} 
\usepackage{enumitem}
\usepackage{diagbox}
\usepackage{amsmath}
\usepackage{xspace}
\usepackage{soul}
\usepackage{media9}

\usepackage[ruled]{algorithm2e} 

\SetAlFnt{\small}
\SetAlCapFnt{\small}
\SetAlCapNameFnt{\small}
\SetAlCapHSkip{0pt}

\newcommand{\sysname}{MTV-Inpaint\xspace}

\setcopyright{none}
\renewcommand\footnotetextcopyrightpermission[1]{} 
\begin{document}
\title{\sysname: Multi-Task Long Video Inpainting}



\author{Shiyuan Yang}
\affiliation{%
  \institution{City University of Hong Kong and Tianjin University}
  \country{China}
}

\author{Zheng Gu}
\affiliation{%
  \institution{Shenzhen University}
  \country{China}}

\author{Liang Hou}
\affiliation{%
  \institution{Kuaishou Technology}
  \country{China}}

\author{Xin Tao}
\affiliation{%
  \institution{Kuaishou Technology}
  \country{China}}

\author{Pengfei Wan}
\affiliation{%
  \institution{Kuaishou Technology}
  \country{China}}

\author{Xiaodong Chen}
\affiliation{%
  \institution{Tianjin University}
  \country{China}}

\author{Jing Liao}
\affiliation{%
\institution{City University of Hong Kong}
  \country{China}}

\renewcommand{\shorttitle}{Yang et al. \sysname}

\begin{abstract}
Video inpainting involves modifying local regions within a video, ensuring spatial and temporal consistency. Most existing methods focus primarily on scene completion (\ie, filling missing regions) and lack the capability to insert new objects into a scene in a controllable manner. 
Fortunately, recent advancements in text-to-video (T2V) diffusion models pave the way for text-guided video inpainting. However, directly adapting T2V models for inpainting remains limited in unifying completion and insertion tasks, lacks input controllability, and struggles with long videos, thereby restricting their applicability and flexibility.
To address these challenges, we propose \sysname, a multi-task video inpainting framework capable of handling both traditional scene completion and novel object insertion tasks. To unify these distinct tasks, we design a dual-branch spatial attention mechanism in the T2V diffusion U-Net, enabling seamless integration of scene completion and object insertion within a single framework. In addition to textual guidance, \sysname supports multimodal control by integrating various image inpainting models through our proposed image-to-video (I2V) inpainting mode. Additionally, we propose a two-stage pipeline that combines keyframe inpainting with in-between frame propagation, enabling \sysname to effectively handle long videos with hundreds of frames.
Extensive experiments demonstrate that \sysname achieves state-of-the-art performance in both scene completion and object insertion tasks. Furthermore, it demonstrates versatility in derived applications such as multi-modal inpainting, object editing, removal, image object brush, and the ability to handle long videos. Project page: 
\textcolor{red}{\url{https://mtv-inpaint.github.io/}.}

\end{abstract}

%
%


\graphicspath{
{figs/}
}

%
%

\newcommand{\etal}{\emph{et al.}}
\newcommand{\ie}{\emph{i.e.}}
\newcommand{\eg}{\emph{e.g.}}

\begin{teaserfigure}
  \includegraphics[width=\textwidth]{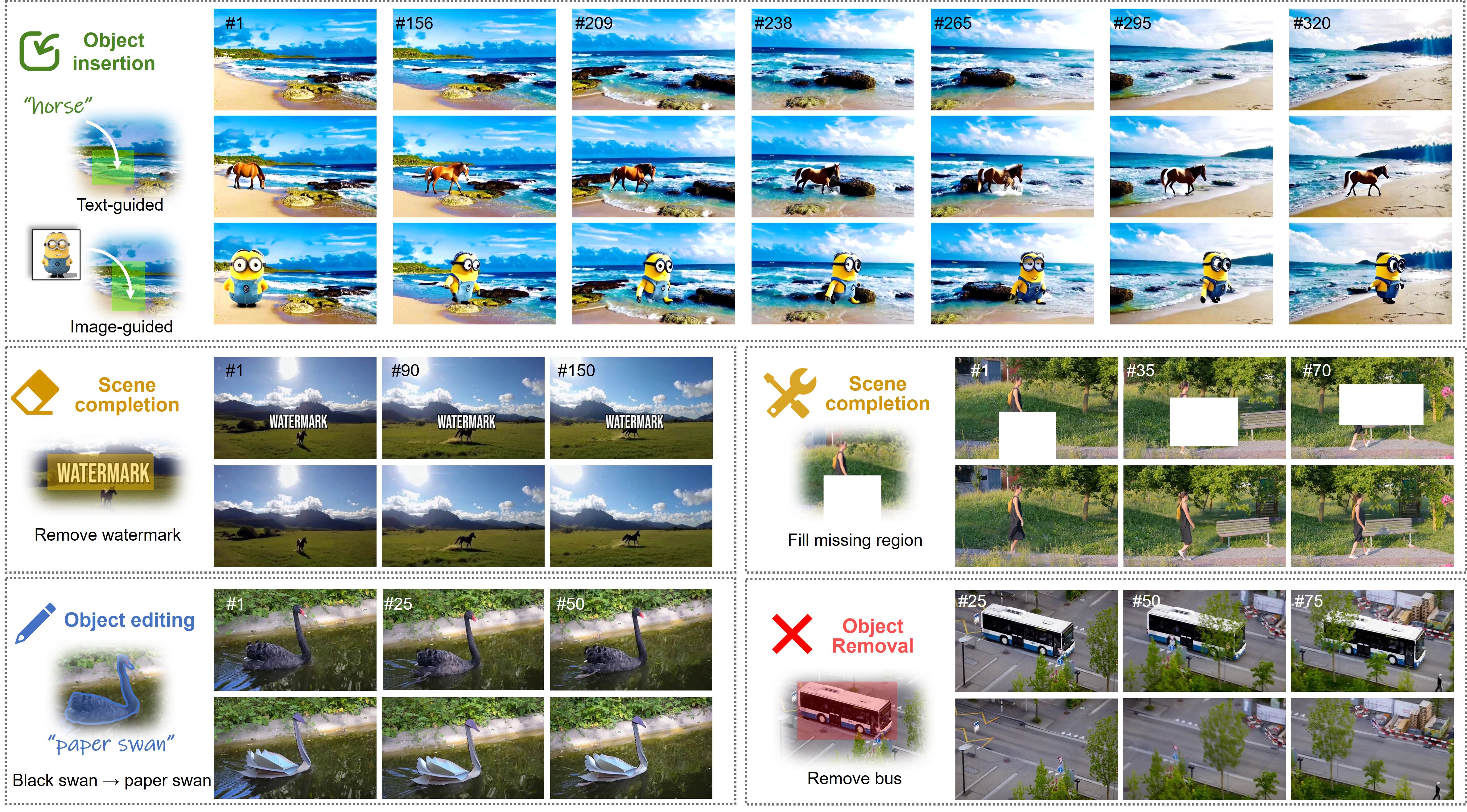}
  \caption{\sysname is a unified video inpainting framework that supports multiple tasks, such as text/image-guided object insertion, scene completion, and derived applications like object editing and removal. It is also capable of handling long videos with hundreds of frames.
  }
  \label{fig.teaser}
\end{teaserfigure}

\maketitle


\section{Introduction} \label{sec.intro}
Video inpainting refers to the process of altering the static or dynamic localized regions within a video, ensuring that the inpainted video exhibits smooth and natural transitions in both spatial and temporal dimensions.

Most existing video inpainting approaches primarily address the problem of unconditional scene completion \cite{e2fgvi, propainter, green}, which involves filling the target regions in a video (\eg, watermark removal) without user guidance. However, these methods lack the ability to perform user-guided object insertion, which entails adding new objects into a scene in a controllable manner.

Fortunately, recent advancements in text-to-video (T2V) conditional generative diffusion models \cite{videoldm,animatediff,sora} have made it possible to inpaint objects into videos under user's guidance. However, current research on text-guided object insertion remains limited. To the best of our knowledge, only CoCoCo~\cite{cococo} has achieved reasonable text-guided object insertion, yet its controllability is limited compared to multimodal conditions commonly employed in the image inpainting area. Additionally, it lacks adaptation for traditional video completion tasks and only supports handling short video clips with limited frames, further limiting its practical use.

%
In practice, users often desire a versatile inpainting solution that is capable of handling both classical scene completion task, as well as the novel object insertion task with customized motion trajectories.
Moreover, the inserted object may not solely be described by text prompt alone; users may prefer to include additional input conditions, such as an exemplar image describing the specific appearance of the object, in case they have more customized needs, as shown in the third row in Figure~\ref{fig.teaser}.
However, up until now, there is no such comprehensive solution offered to this issue.

To address the aforementioned objective, we aim to leverage T2V generative models as the foundation, with the following challenges to be addressed: 
(1) \textbf{Task unification problem}, scene completion and object insertion are inherently distinct tasks with different goals and requirements. How can we unify them into a single framework? 
(2) \textbf{Controllability problem}, for the object insertion task, relying solely on text prompts is insufficient for fine-grained control. How can we incorporate diverse input conditions to enhance the controllability?
(3) \textbf{Long video problem}, existing T2V models are trained to generate limited length of video. How can we extend these models to support longer videos? 

In this work, we propose \sysname, a multi-task video inpainting framework built upon T2V diffusion model. Other than scene completion, our method further allows users to insert an object with more forms of control and customized trajectory in long videos. 

Firstly, unifying object insertion and scene completion is not trivial, as these tasks are fundamentally different in nature. Object insertion requires generating a temporally consistent object within the masked region, ensuring the object's coherence across frames. In contrast, scene completion focuses on filling the masked region based on the surrounding context, where the inpainted content may dynamically vary over time.  As such, we introduce dual-branch spatial block with shared temporal block in the T2V diffusion U-Net, where one spatial branch is tailored  for object insertion and the other is dedicated to scene completion.

Secondly, to enhance controllability with more diverse conditioning beyond T2V, a straightforward approach is to train multiple adapters tailored for different conditions.  However, this requires designing modality-specific architectures and training on different datasets, which can be resource-intensive and challenging. Fortunately, we notice that the image inpainting domain already features models capable of flexible conditional control, including text, exemplar image, edge maps etc. This inspires us to leverage the strengths of these off-the-shelf methods for video inpainting. Specifically, we integrate I2V (image-to-video) inpainting mode into our method, which bridges video inpainting with existing image inpainting tools. In this mode, any third-party image inpainting method can be used to inpaint the first frame, which is then propagated across subsequent frames.

Finally, for the long video inpainting task, current T2V models, which are pretrained to generate short video clips, struggle to maintain quality when applied directly to longer videos. To address this, we propose a two-stage pipeline: keyframe inpainting followed by in-between inpainting. We first inpaint keyframes distributed across the original video, leveraging either the T2V or I2V inpainting modes, then we iteratively fill the intermediate frames between each pair of adjacent inpainted keyframes. This process, which we term K2V (keyframe-to-video) inpainting, ensures smooth temporal transitions and consistent inpainting across the entire video.

We evaluated our method on two primary video inpainting tasks: text-guided object insertion and scene completion. The results demonstrate state-of-the-art performance compared to existing baselines. Additionally, we highlight the versatility of our method through its application to derived tasks, including multi-modal inpainting, object removal, editing, and image object brush, showcasing its broad utility as a video inpainting tool. 

In summary, our contributions are as follows:
\begin{itemize}[leftmargin=*]
    \item We propose a multi-task video inpainting framework capable of handling both object insertion and scene completion tasks, as well as derived tasks like object removal and editing.
    \item Our framework brings more controllability for video inpainting by connecting with any existing powerful image inpainting tool via I2V inpainting mode.
    \item We design a two-stage pipeline consists of keyframe + in-between inpainting, to support inpainting for longer videos while ensuring temporal coherence.
\end{itemize} 

\section{Related Work} \label{sec.rw}

\subsection{Image/Video Generation} \label{sec.rw.1}

Text-to-image (T2I) diffusion models, such as Stable Diffusion-series \cite{ldm}, Flux-series~\cite{flux}, have revolutionized the field of image synthesis. Building on these foundation models, various works have significantly enhanced the controllability of image generation by incorporating additional modules that handle diverse conditions, such as sketch maps~\cite{t2iadapter}, skeletons~\cite{controlnet}, bounding boxes~\cite{gligen}, and reference images~\cite{ipadapter} etc. These advancements have greatly improved the flexibility and applicability of T2I models.

The success of T2I models has also propelled advancements in text-to-video and image-to-video generation. Video generation models are often extended from image models by adding temporal layers~\cite{animatediff,makeavideo} to ensure temporal consistency across frames. Prominent T2V models, including SORA~\cite{sora}, Gen-series~\cite{gen1}, and CogVideo-X~\cite{cogvideox}, are capable of generating high-quality videos from textual descriptions. Similarly, notable I2V models such as Dynamicrafter~\cite{dynamicrafter} and Stable Video Diffusion~\cite{svd} focus on animating static images into dynamic video clips.
Building on these foundational models, researchers have explored greater controllability in video generation, including customizing subject identity~\cite{animateanyone, dreamvideo}, motion~\cite{magicdance}, and camera movement~\cite{motionctrl}, enabling more flexible video creation workflows.

\subsection{Image/Video Inpainting}\label{sec.rw.2}

As a promising generative paradigm, T2I diffusion models have been successfully applied to image inpainting. Various T2I-based approaches have emerged, including unconditional~\cite{repaint}, text-driven~\cite{ldm, hdpainter, brushnet}, image-driven~\cite{pbe, anydoor, realfill}, shape-driven~\cite{smartbrush}, instruction-driven~\cite{instinpaint}, multi-task~\cite{powerpaint}, multi-view~\cite{mvinpainter}, and multi-modal inpainting~\cite{pilot, unipaint}, showcasing their versatility across diverse inpainting tasks.

Similarly, T2V diffusion models have shown great potential in video inpainting. Unlike traditional video inpainting models~\cite{dfgvi, fegvc, ecfvi, e2fgvi, propainter}, which are limited to completing missing regions in videos without user guidance, T2V models can incorporate text prompts to guide the inpainting process, enabling new applications such as object insertion. However, research on leveraging T2V diffusion models for object inpainting remains limited.
For example, works like \citet{floed} and \citet{green} leverage T2V diffusion models for consistent video completion, yet they are still unable to insert new objects into the scene. VideoComposer~\cite{videocomposer} trained T2V model on static masks, limiting its ability to inpaint moving objects. AVID~\cite{avid} uses ControlNet-like adapters in text-driven inpainting, which requires source structures, thereby limiting its ability to add new objects as the original structures are often unavailable.
VideoPoet~\cite{videopoet} and Lumiere~\cite{lumiere} demonstrate the ability to add objects within a static mask, but using static mask cannot specify object movements.
CoCoCo~\cite{cococo} improves dynamic object insertion by training on object-aware masks. However, this biases the model to generate objects within the masked region, making it more challenging to adapt to scene completion.
In contrast, our method addresses this limitation with a dual-branch architecture that unifies object insertion and scene completion in a single model.
%
Furthermore, the input conditions for prior methods are typically limited to text guidance, reducing their controllability. Our approach supports I2V inpainting mode, enabling integration with various image inpainting methods for enhanced controllability. Additionally, our K2V inpainting mode extends diffusion models to handle longer videos with consistent temporal coherence. These advancements make our approach more practical and versatile for real-world applications.
\section{Method} \label{sec.meth}

\begin{figure*}[t]
    \centering
    \includegraphics[width=1\linewidth]{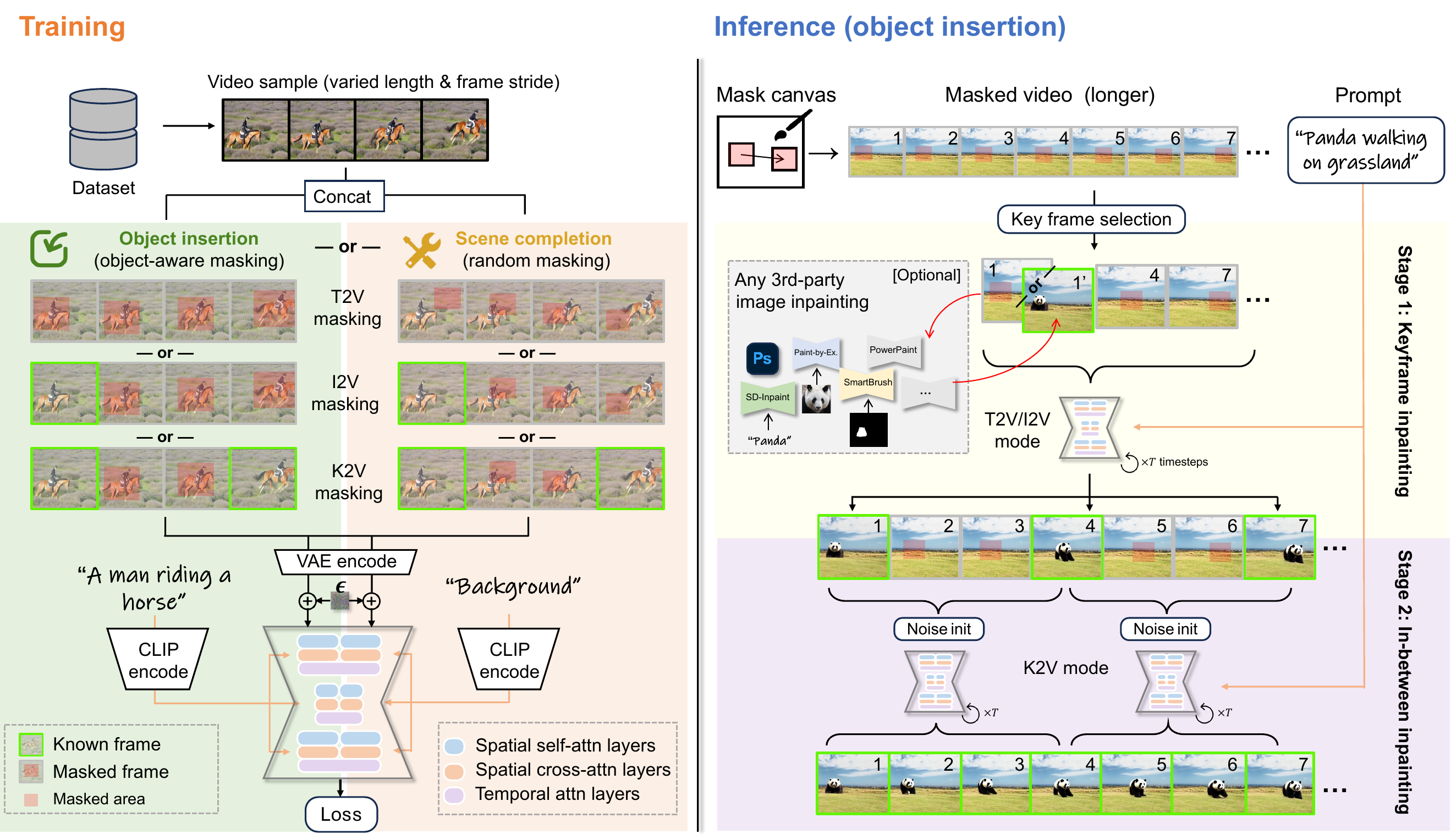}
    \caption{Our VideoPaint framework. During training, we train object insertion and scene completion tasks with dual-branch U-Net, using object-aware masks and random masks respectively. Concurrently, we employ three frame masking modes: text-to-video(T2V), image-to-video (I2V), and keyframe-to-video (K2V). During the inference, our method can perform basic T2V inpainting, or I2V inpainting, given that the first frame is obtained from 3rd party image inpainting tool. To handle longer video, we first use T2V/I2V mode to inpaint keyframes, then use K2V mode to inpaint remaining in-between frames.}
    \label{fig.pipeline}
\end{figure*}

\subsection{Overview} \label{sec.meth.overview}

\subsubsection{Task Formulation} 

Given an original source video $\mathbf{X}^{1:N}$ with $N$ frames, the user should also provide its binary mask sequence $\mathbf{M}^{1:N}$, where a value of 1 indicates the regions to be inpainted. To simplify the mask generation process, we let users draw bounding boxes on the first frame, the last frame, and optionally on certain intermediate frames, as well as specify a trajectory path connecting these boxes. These are then used to interpolate an $N$-box sequence, from which we obtain the final box-shaped mask sequence $\mathbf{M}^{1:N}$.
\begin{itemize}[leftmargin=*]
\item For object insertion task,  the masks specify the spatial-temporal regions where the object is expected to appear, serving as a signal for the object’s motion. In addition, the user must provide a text prompt describing the desired object and may optionally supply an inpainted first frame to further define the object’s initial appearance.
\item For scene completion task, the masks indicate the regions that require refilling. In this case, a text prompt is not needed as the filling content will be automatically determined by surrounding context. Also, users may optionally provide an inpainted first frame to predefine the initial desired content. 
\end{itemize}

\subsubsection{Overall Pipeline} 

Our overall pipeline is illustrated in Figure.~\ref{fig.pipeline}. During training, we employ dual spatial branch U-Net to handle both object insertion and scene completion tasks. For object insertion, we train the model using object-aware masks, while for scene completion, we use random masks. Simultaneously, we train the U-Net with three different frame masking modes: (1) Text-to-Video (T2V) mode: inpaints all frames by text prompt. (2) Image-to-Video (I2V) mode: inpaints subsequent frames based on a provided first frame and text prompt. (3) Keyframe-to-Video (K2V) mode: inpaints in-between frames based on two keyframes given at the beginning and end of a sequence.
During inference, our method supports various inpainting scenarios. It can perform basic T2V inpainting, or I2V inpainting, where the first frame is provided by a third-party image inpainting tool. For longer videos, we first inpaint keyframes using T2V or I2V modes, and then use the K2V mode to inpaint the remaining intermediate frames, as shown in the right side of Figure~\ref{fig.pipeline}. This two-step pipeline ensures temporal consistency across the entire video.

\subsection{Model Architecture} \label{sec.meth.model}

Inspired by the image inpainting diffusion model ~\cite{ldm}, we implemented our method as a latent 3D diffusion U-Net with masking conditions, \ie, the inputs are first encoded into the latent space by variational auto-encoder (VAE) before feeding into the model, which is a triplet consisting of a noised video latent $\mathbf{x}_t$, a down-sampled binary mask sequence $\mathbf{m}$, and a masked video latent $\mathbf{x}_m = \mathbf{x} \odot \mathbf{m}$,  concatenated along the channel dimension. The U-Net (parameterized by $\boldsymbol{\theta}$) also considers the timestep $ t $ and a text prompt embedding $ \mathbf{c} $ as conditions and predicts the noise $ \boldsymbol{\epsilon}_\theta $. The model is optimized using the following masked denoising loss:
\begin{equation}   \label{eq.loss}
\mathcal{L}=\mathbb{E}_{\mathbf{x}, \mathbf{m},\mathbf{x}_m, \mathbf{c}, t, \boldsymbol{\epsilon} }\left[\left\|\left(\boldsymbol{\epsilon}-\boldsymbol{\epsilon_\theta}\left(\mathbf{x}_t, \mathbf{m},\mathbf{x}_m \mathbf{c}, t \right)\right)\
\odot (\lambda\mathbf{m}+\mathbf{1})   \right\|_2^2\right],
\end{equation}
where $ \mathbf{x}_t = \alpha_t \mathbf{x} + \sigma_t \boldsymbol{\epsilon} $,  $ \boldsymbol{\epsilon}\sim \mathcal{N}(\mathbf{0}, \mathbf{I}) $, $ \alpha_t $ and $ \sigma_t $ are time-dependent DDPM hyper-parameters~\cite{ddpm}. $\lambda$ is the loss weight for masked area.

In terms of model architecture, our method incorporates the following improvements: To handle both object insertion and scene completion tasks simultaneously, we introduce dual-branch spatial attention in the U-Net. Each branch consists of dual reference self-attention and cross-attention blocks, enabling the model to specialize in distinct inpainting requirements for the two tasks.

\subsubsection{Dual-Branch Spatial Attention} 

Object insertion task and the scene completion task are fundamentally different in their requirements for the inpainted content, and are sometimes even contradictory. Object insertion requires generating a consistent and coherent object within the masked region across all frames, while scene completion focuses on filling the masked region based on the surrounding context. Due to the motion of foregrounds, backgrounds, and the mask itself, the synthesized content for scene completion may vary across frames. Unifying these two tasks within a single model is therefore challenging. 

Previous studies have shown that, in diffusion U-Net, spatial attention primarily handles object synthesis: where spatial self-attention layers determine fine-grained appearance~\cite{masactrl}, spatial cross-attention layers fuse external textual conditions and control the semantic layout~\cite{p2p}. Temporal attention, on the other hand, is responsible for maintaining temporal coherence across frames ~\cite{animatediff}. Considering that the primary difference between object insertion and scene completion lies in their content generation objectives, while both require temporal consistency. Therefore, we designed the U-Net architecture with dual-branch spatial attention with shared temporal attention. One spatial branch is specialized for object insertion, and the other is tailored for scene completion, as illustrated in left side of Figure.~\ref{fig.pipeline}.

\subsubsection{Dual-Reference Self-Attention} 
As stated above, self-attention plays a critical role in controlling spatial appearance during the generation process. Considering that our method is designed to support not only T2V inpainting but also I2V and K2V inpainting modes (introduced in the following sections), it becomes essential to incorporate information from the first and last frames during generation. There have been numerous studies employing reference attention~\cite{masactrl,tokenflow,tav,magicanimate,stylealign}, demonstrating its effectiveness in preserving object identity.
Similarly, we also extend the native single-frame self-attention mechanism to dual-reference self-attention, where each frame attends not only to itself but also to the first and last frames. This mechanism is formulated as follows:

\begin{equation}   \label{eq.selfattn}
\text{SelfAttn}(\mathbf{Q}^i, \mathbf{K}^i, \mathbf{V}^i) = 
\text{softmax}\left(\frac{\mathbf{Q}^i \left[\mathbf{K}^i; \mathbf{K}^1; \mathbf{K}^n\right]^\top}{\sqrt{d}}\right) 
\cdot \left[\mathbf{V}^i; \mathbf{V}^1; \mathbf{V}^n\right],
\end{equation}   
where $\mathbf{Q}^i$, $\mathbf{K}^i$, and $\mathbf{V}^i$ represent the query, key, and value feature of frame $i$, and $\mathbf{K}^1$, $\mathbf{K}^n$, $\mathbf{V}^1$, $\mathbf{V}^n$ represent the key and value of the first and last frames, respectively, $d$ is the feature dimension. The operation $[;]$ denotes concatenation along the feature dimension.

\subsection{Training-Time Masking Scheme} \label{sec.meth.masking}
Our training-time masking scheme comprises two components: the regional masking scheme, tailored for training different tasks, and the frame masking scheme, designed to train different conditioning modes.

\subsubsection{Regional Masking Scheme} 
Given the different generation objectives, we employ distinct regional masking schemes for different tasks.
For object insertion, using object-aware masks is essential to ensure accurate synthesis of the object within the masked region. To achieve this, we train the model using object-centered videos, masks, and associated prompts sourced from object tracking and segmentation datasets. During training, we always dilate the object masks into bounding box masks to align with the box masks used during inference.
For scene completion, we train the model with randomly generated masks. This approach is designed to cover diverse scenarios encountered during inference. Since the inpainted content is typically non-deterministic and often involves background regions, we empirically fix the prompt to 'background'. This setup was found to perform well, even in cases where the masked regions included parts of the foreground.

\subsubsection{Frame Masking Scheme} 
To enable our framework to handle diverse inpainting modes and downstream applications, we train the model with three distinct frame masking schemes:

\begin{itemize}[leftmargin=*]
    \item T2V (Text-to-Video) mode: All frames are masked, and the model inpaints the entire video based on a text prompt.
    \item I2V (Image-to-Video) mode: The first frame is unmasked, and the model inpaints subsequent frames based on the first frame and a text prompt.
    \item K2V (Keyframe-to-Video) mode: The first and last frames are unmasked, and the model inpaints the intermediate frames based on these two keyframes and a text prompt.
\end{itemize}

The T2V mode trains the model's inherent T2V inpainting capability. The I2V mode enables the framework to utilize outputs from third-party image inpainting models, providing enhanced controllability. The K2V mode is primarily designed to complement the T2V or I2V modes during inference, facilitating long video inpainting as described in the following section.

\subsection{Inference-Time Long Video Inpainting } \label{sec.meth.long}

In practice, users often provide long videos, typically exceeding the video length used during training. To adapt our model for long video inpainting, we propose a two-stage long video inpainting pipeline.

\subsubsection{Two-Stage Inference Pipeline} 
The two-stage inference pipeline consists of keyframe inpainting followed by in-between inpainting, as illustrated on the right side of Figure~\ref{fig.pipeline}. While such keyframe-based in-between techniques have been adopted in long video processing works~\cite{tokenflow, tcc}, our approach focuses on generation rather than propagation from the keyframes.
In the keyframe inpainting stage, we sample keyframes 
$\{\mathbf{x}^{k_1}, \mathbf{x}^{k_2}, \dots, \mathbf{x}^{k_n}\}$ 
from the source video, where $k_1 = 1$ and $k_n = N$, ensuring that both the first and last frames are selected as keyframes. These keyframes are then inpainted using either the T2V or I2V inpainting mode, resulting in the inpainted keyframes 
$\{\hat{\mathbf{x}}^{k_1}, \hat{\mathbf{x}}^{k_2}, \dots, \hat{\mathbf{x}}^{k_n}\}$.
In the in-between inpainting stage, the K2V inpainting mode is repeatedly applied to the intermediate frames in each interval enclosed by two adjacent inpainted keyframes $\hat{\mathbf{x}}^{k_i}$ and $\hat{\mathbf{x}}^{k_{i+1}}$. By iteratively applying this approach across all intervals, we are able to inpaint a long video while maintaining temporal consistency.

\subsubsection{K2V Prior Noise Initialization} \label{meth.long.noise_init}
Existing research~\cite{flawnoise} has identified a noise gap between diffusion training and inference. During training, the model learns to denoise a noisy input where the signal is not completely destroyed, whereas in the test stage, samples are generated from pure random noise, such discrepancy can sometimes lead to temporal sudden changes.
To address this issue in our K2V mode, inspired by previous noise initialization techniques~\cite{controlavideo, preserve, noisewarp, freenoise}, we propose K2V prior noise initialization. Instead of sampling from random noise, we initialize the noise by utilizing prior information from the known first and last frames.


Formally, let $\mathbf{x}^{k_1:k_2} = \{\hat{\mathbf{x}}^{k_1}, \mathbf{x}^{k_1+1}, \dots, \mathbf{x}^{k_2-1}, \hat{\mathbf{x}}^{k_2}\}$ denote a sequence of interval frames bounded by two known adjacent keyframes $\hat{\mathbf{x}}^{k_1}$ and $\hat{\mathbf{x}}^{k_2}$. We define $\mathcal{P}(\mathbf{x}^i, \mathbf{m}^i)$ as the operation that extracts and resizes the local regions of $\mathbf{x}^i$ specified by $\mathbf{m}^i$ to the median size.
First, the local target regions of the intermediate frames are determined through linear interpolation in the extracted regions:
$
\mathcal{P}(\hat{\mathbf{x}}^i, \mathbf{m}^i) = (1 - \eta) \mathcal{P}(\hat{\mathbf{x}}^{k_1}, \mathbf{m}^i) + \eta \mathcal{P}(\hat{\mathbf{x}}^{k_2}, \mathbf{m}^i), \quad \eta = \frac{i - k_1}{k_2 - k_1}.
$
The interpolated regions, $\mathcal{P}(\hat{\mathbf{x}}^i, \mathbf{m}^i)$, are then resized and pasted back into the corresponding regions of $\mathbf{x}^i$ to produce the updated frame $\hat{\mathbf{x}}^i$.
Next, we perform one-step DDPM forward noise addition at timestep $\tau$ to generate a noisy sequence:
$\hat{\mathbf{x}}_\tau^{k_1:k_2} = \sqrt{\bar{\alpha}_\tau} \hat{\mathbf{x}}^{k_1:k_2} + \sqrt{1 - \bar{\alpha}_\tau} \boldsymbol{\epsilon}^{k_1:k_2}$.
Finally, we combine the low-frequency component of $\hat{\mathbf{x}}_\tau^{k_1:k_2}$ with the high-frequency component of $\boldsymbol{\epsilon}^{k_1:k_2}$ using a Gaussian low-pass filter in the frequency domain:

\begin{equation}   \label{eq.noise_init}
\hat{\boldsymbol{\epsilon}}^{k_1:k_2} = \mathcal{F}^{-1} \big[\mathcal{F}(\hat{\mathbf{x}}_\tau^{k_1:k_2}) \odot G_{\text{LPF}} + \mathcal{F}(\boldsymbol{\epsilon}^{k_1:k_2}) \odot (1 - G_{\text{LPF}})\big],
\end{equation}
where $\mathcal{F}$ and $\mathcal{F}^{-1}$ denote the 3D Fourier transform and its inverse operation, respectively, and $G_{\text{LPF}}$ is a Gaussian low-pass filter in the frequency domain.
The new initialized noise $\hat{\boldsymbol{\epsilon}}^{k_1:k_2}$ incorporates prior information transitioning from $\hat{\mathbf{x}}^{k_1}$ to $\hat{\mathbf{x}}^{k_2}$. Our ablation study demonstrates that this approach stabilizes the generated results, leading to smoother transitions and increased temporal consistency.




\section{Experiments} \label{sec.exp}

\subsection{Experimental Setup} \label{sec.exp.setup}

\paragraph{Implementation Details}  \label{sec.exp.setup.impl}
Our inpainting model is finetuned from a text-to-video diffusion U-Net~\cite{modelscope}, with 5 additional zero-initialized input channels to encode masking conditions. To enable the model to inpaint videos of varied lengths, motion scales and sizes, the training video clips are sampled with dynamic lengths ranging from 8 to 24 frames, and dynamic frame strides ranging from 1 to 10, as well as dynamic resolution of $320\times512$ or $512\times512$ or $512\times320$. The loss weight of masked region $\lambda$ in Eq.~\ref{eq.selfattn} is set to 2. The T2V, I2V, and K2V frame masking modes are applied with equal probabilities of $1/3$.  During inference, we use the DDIM sampler~\cite{ddim} with 30 steps and a classifier-free guidance scale of 8 ~\cite{cfg}. 

\paragraph{Datasets}  \label{sec.exp.setup.datasets}
For training, we use object tracking and segmentation datasets, including YoutubeVOS~\cite{vos}, YoutubeVIS~\cite{vis}, MOSE~\cite{mose}, VIPSeg~\cite{vipseg}, UVO~\cite{uvo}, and GOT~\cite{got}, comprising approximately 20k videos in total. For evaluation, we utilize videos and masks from DAVIS~\cite{davis} as well as some self-collected data. Additionally, we use ChatGPT-4o~\cite{gpt4} to generate multiple prompts for each masked video, resulting in 220 video-prompt pairs.

\paragraph{Metrics}  \label{sec.exp.setup.metrics}
For the object insertion task, to evaluate object-prompt alignment, we calculate the regional CLIP image-text score (CLIP-T) within the masked area~\cite{clipscore}. To measure object temporal consistency (TempCons), we compute the regional CLIP image-image score between every pair of consecutive frames, following~\cite{gen1}. To assess object insertion spatial accuracy, we use GroundingDINO~\cite{groundingdino} to detect object bounding boxes in the generated videos, which are then compared to the input masks to calculate the mean intersection over union (mIoU). Additionally, we employ the ImageReward~\cite{imagereward} model to evaluate the overall visual and aesthetic quality, as it has been shown to align closely with human judgment.
For the scene completion task, we use PSNR and LPIPS~\cite{lpips} to evaluate completion quality at both the pixel and feature levels. We also assess full video temporal consistency and overall visual quality using TempCons and ImageReward as described previously.

\paragraph{Baselines}  \label{sec.exp.setup.baseline}
For the object insertion task, we compare our method with the most recent CoCoCo~\cite{cococo}. Given the limited research focus on text-to-video object insertion, we also implement two additional baselines: Zeroscope-blend and Animate-inpaint.
Zeroscope-blend is a combination of the Zeroscope~\cite{modelscope} T2V model and the latent blending technique~\cite{bld}. Specifically, we adapt Zeroscope for zero-shot text-guided inpainting by blending the known regions into the denoised latent representation at each timestep.
Animate-inpaint integrates AnimateDiff~\cite{animatediff} with the SD-inpaint model~\cite{ldm}. In this approach, we insert temporal layers into the SD image inpainting model and fine-tune these layers to adapt the model for video inpainting tasks.
For the scene completion task, we compare our method with CoCoCo, and two state-of-the-art non-diffusion-based models: E2FGVI\cite{e2fgvi} and ProPainter~\cite{propainter}, both of which are dedicated to video completion.

\subsection{Qualitative Results} \label{sec.exp.ql}

\subsubsection{Object Insertion} \label{sec.exp.ql.obj_ins}

\begin{figure*}
    \centering
    \includegraphics[width=1\linewidth]{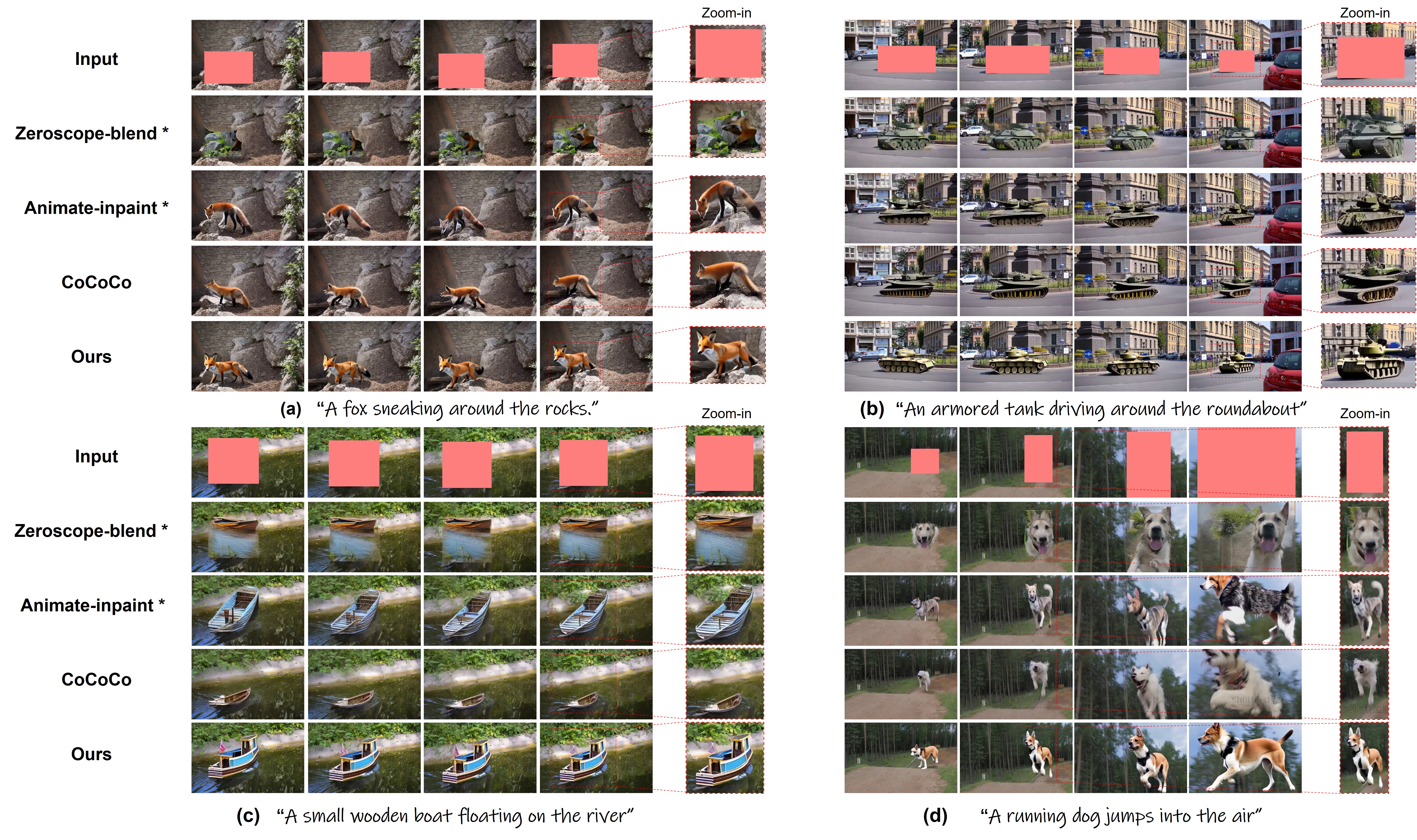}
    \caption{Quantitative comparison for object insertion evaluation. We recommend watching our supplementary video for dynamic results. Methods marked with an asterisk are not existing works but have been implemented by us.}
    \label{fig.obj_ins}
\end{figure*}

We present a side-by-side visual comparison with baseline methods in Figure~\ref{fig.obj_ins}. 
%
%
While Zeroscope-blend sometimes is capable of synthesizing the desired objects within the target regions, the transitions between the filled regions and the background are often visually inconsistent (\eg, see Figure~\ref{fig.obj_ins}(c) boat examples). This issue stems from the latent blending strategy, which forcibly merges the background into the target regions, leading to unharmonious edges.
In contrast, Animate-inpaint generates more natural transitions compared to Zeroscope-blend, as it has been fine-tuned on video data. However, it falls short in preserving object identity (\eg, the dog in Figure~\ref{fig.obj_ins}(d) exhibits noticeable variations). This limitation arises from using an image-based model as its foundation; even with the addition of temporal layers, its temporal consistency remains weaker than that of a fully pre-trained video model. 
CoCoCo achieves relatively reasonable results; however, our inpainting results exhibit superior visual quality, particularly under scenarios involving large motion (\eg, the jumping dog in Figure~\ref{fig.obj_ins}(d)). This improvement can be attributed to our training strategy, where we intentionally use frame stride with greater variation to ensure the model is better adapted to handling larger motions, especially for keyframe inpainting tasks.

\subsubsection{Scene Completion} \label{sec.exp.ql.scn_comp}
\begin{figure*}
    \centering
    \includegraphics[width=1\linewidth]{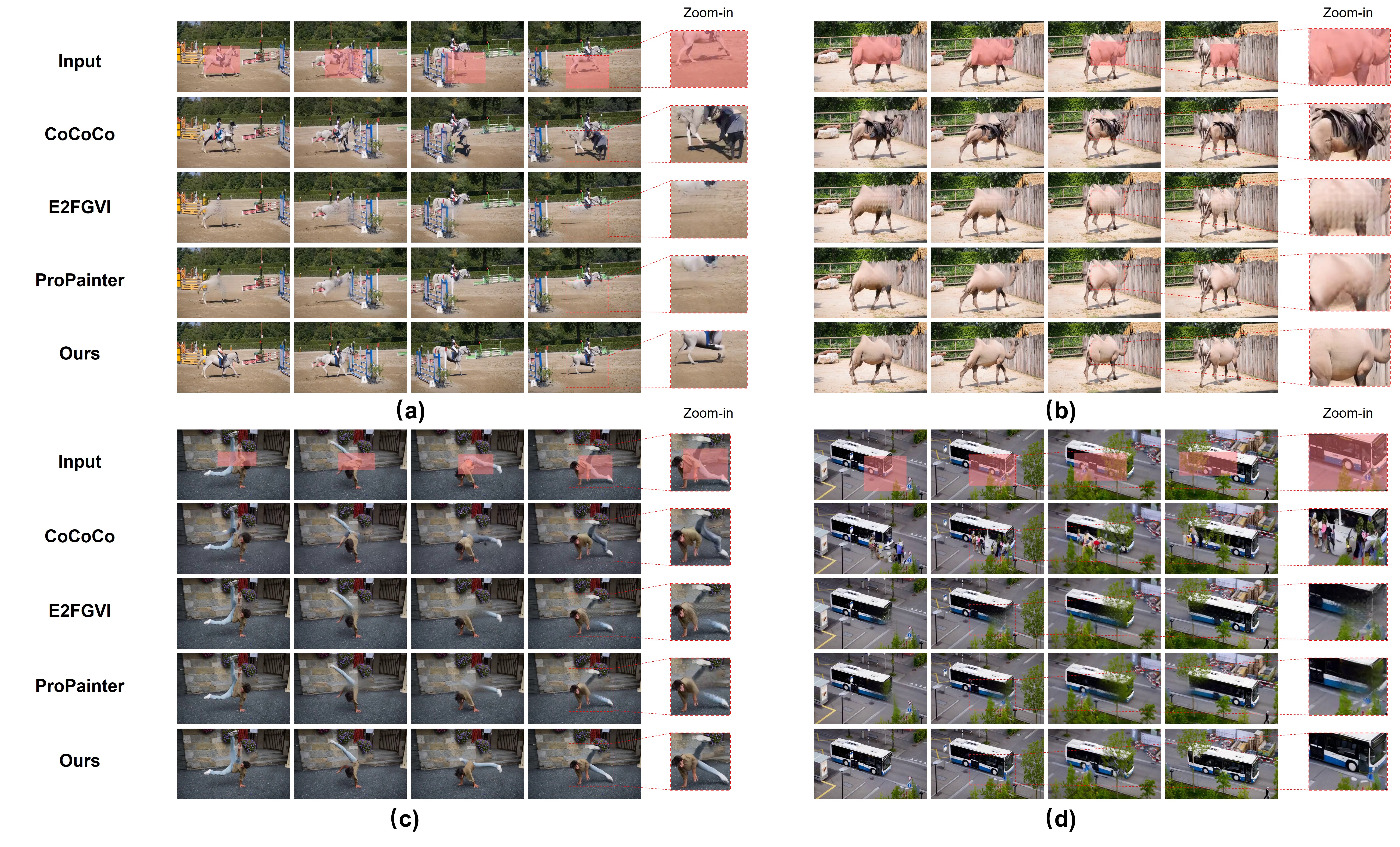}
    \caption{Quantitative comparison for scene completion evaluation. We recommend watching our supplementary video for dynamic results.}
    \label{fig.scn_comp}
\end{figure*}

Figure~\ref{fig.scn_comp} presents a visual comparison of different methods for the scene completion task. Since CoCoCo was not specifically trained for scene completion, and textual descriptions for the missing regions are hard to define, considering the null condition was paired with diverse general scenes during training, we use the null condition as its input.
From the figure, it is evident that CoCoCo tends to add redundant elements (see zoom-in areas in Figure~\ref{fig.scn_comp}(a), (b) and (d)). This limitation arises from CoCoCo being trained exclusively for object inpainting, this inadvertently brings a side effect to its performance on scene completion task. E2FGVI and ProPainter fall short in handling complex semantics (\eg, failing to inpaint the horse in Figure~\ref{fig.scn_comp}(a)). Additionally, for regions masked across all frames, they often produce blurry results since such information cannot be found elsewhere (see Figure~\ref{fig.scn_comp}(b)). In contrast, our method achieves more plausible results, benefiting from dual-task training as well as the strong priors provided by the T2V diffusion model.

\subsection{Quantitative Results} \label{sec.exp.qt}

\subsubsection{Object Insertion} \label{sec.exp.qt.obj_ins}
Since our baseline CoCoCo supports short video length of 16 frames, for a fair comparison, we also train Animate-Inpaint on 16-frame sequences, this evaluation was conducted on short videos of the same length. 
We report quantitative metrics in comparison with related baselines in Table~\ref{tab.obj_ins}. The results indicate that our method achieves superior performance in text faithfulness (CLIP-T) and temporal consistency (TempCons). Notably, our approach significantly outperforms existing baselines in grounding ability (mIOU) and visual quality (ImageReward), demonstrating state-of-the-art performance.

\begin{table}[h]
    \centering
    \setlength{\tabcolsep}{0.7mm}
    \caption{Quantitative comparison for object insertion evaluation.}
    \begin{tabular}{l|cccc}
        \toprule
        & CLIP-T$\uparrow$ & TempCons$\uparrow$ & mIOU(\%)$\uparrow$ & ImageReward$\uparrow$ \\
        \midrule
        Zeroscope-blend & 27.35 & 93.03 & 41.14 & -0.290 \\
 Animate-Inpaint& 28.01& 93.15& 78.82&-0.099\\
        CoCoCo & 28.53 & 93.16 & 57.63 & 0.026 \\
        \sysname(Ours) & \textbf{28.78} & \textbf{94.82} & \textbf{85.00} & \textbf{0.106} \\
        \bottomrule
    \end{tabular}
    \label{tab.obj_ins}
\end{table}

\subsubsection{Scene Completion} \label{sec.exp.qt.scn_comp}
We report the metrics for the scene completion task in comparison with related baselines in Table~\ref{tab.scn_comp}. In terms of reconstruction, while our PSNR is lower than that of E2FGVI and ProPainter, our LPIPS score is superior. This suggests that our method may not achieve the closest pixel-level match to the ground truth but performs better in feature-level restoration, which aligns more closely with human visual perception. This observation is further supported by our higher ImageReward score, which evaluates overall visual quality. Additionally, our method demonstrates better temporal consistency, making it a competitive tool for such application.

\begin{table}[h]
    \centering
    \setlength{\tabcolsep}{0.7mm}
    \caption{Quantitative comparison for scene completion evaluation.}
    \begin{tabular}{l|cccc}
        \toprule
        & PSNR $\uparrow$ & LPIPS $\downarrow$ & TempCons $\uparrow$ & ImageReward $\uparrow$ \\
        \midrule
        CoCoCo & 23.06 & 0.069 & 93.95 & 0.066 \\
 E2FGVI& 27.88& 0.057& 94.05&0.295\\
        ProPainter & \textbf{28.62}& 0.053 & 94.95 & 0.297 \\
        \sysname(Ours) & 27.52& \textbf{0.043} & \textbf{95.02} & \textbf{0.527} \\
        \bottomrule
    \end{tabular}
    \label{tab.scn_comp}
\end{table}

\subsection{User Study} \label{sec.exp.user_study}
\begin{figure}
    \centering
    \includegraphics[width=1\linewidth]{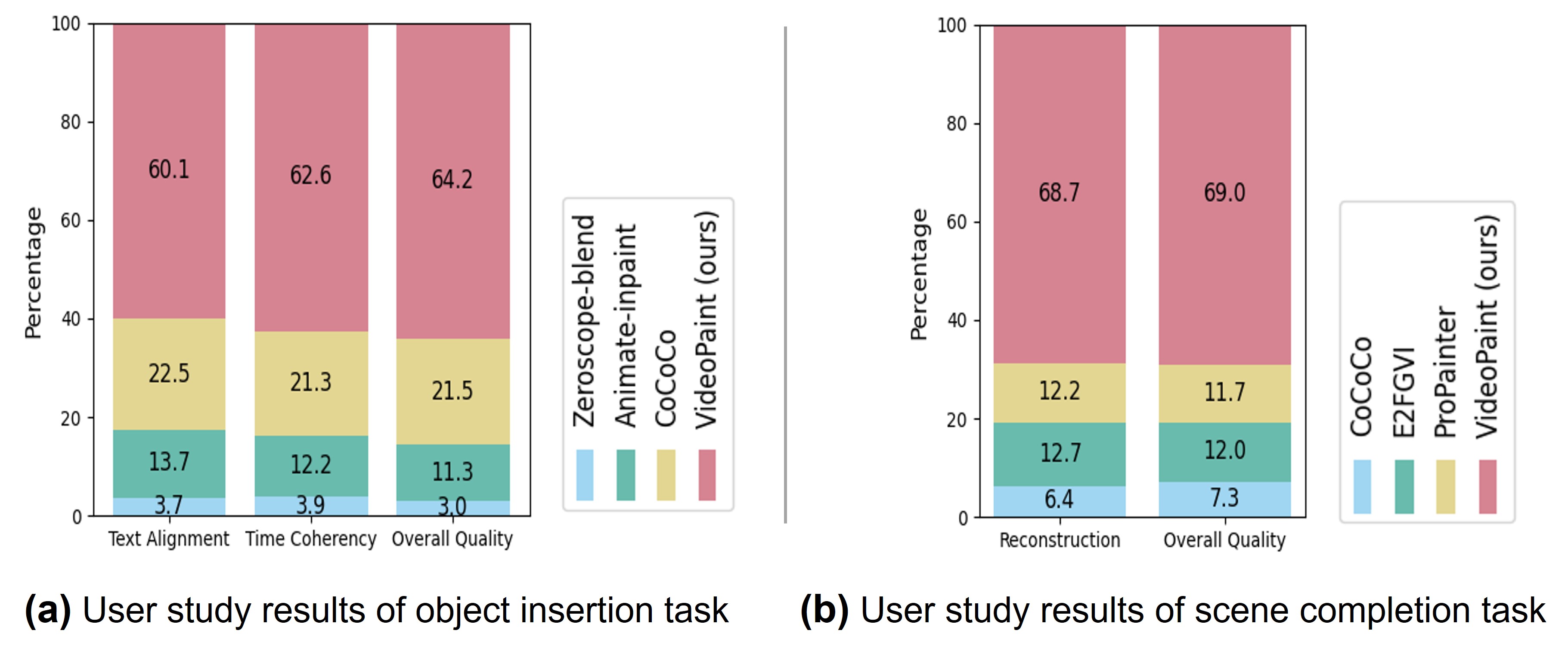}
    \caption{User study results of different methods on \textbf{(a)} object insertion task, and \textbf{(b)} scene completion task.}
    \label{fig.user_study}
\end{figure}

We conducted a user study to evaluate the human perceptual performance of our method. The study included both object insertion and scene completion tasks, comprising a total of 24 questions with 47 participants. To reduce selection bias, the options for each question were randomly shuffled. For the object insertion task, participants were asked to vote for their preferred results based on three criteria: text alignment, temporal coherence, and overall visual quality. For the scene completion task, participants evaluated results based on two criteria: reconstruction ability (with the ground truth provided as a reference) and overall visual quality (reference-free). Figure~\ref{fig.user_study} shows the stacked bar chart of the results. As illustrated, our method achieved the highest preference rate across all tasks and evaluation aspects, demonstrating its superior perceptual quality.

\begin{figure*}
    \centering
    \includegraphics[width=1\linewidth]{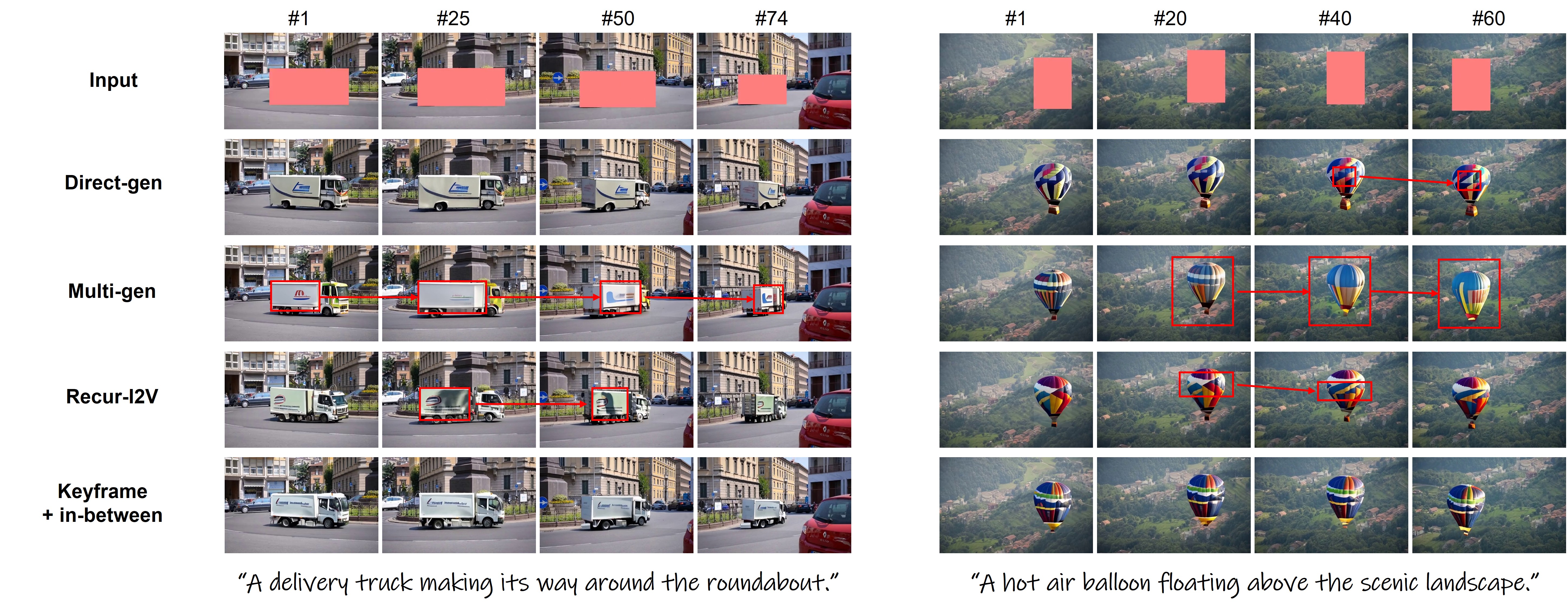}
    \caption{Visual examples from different long video generation strategies.}
    \label{fig.long_example}
\end{figure*}

\subsection{Ablation Study} \label{sec.exp.abl}
We conducted several ablation studies to validate the effectiveness of the proposed components in our method.

\subsubsection{Necessity of the Dual-Branch Design}
\begin{figure}
    \centering
    \includegraphics[width=0.9\linewidth]{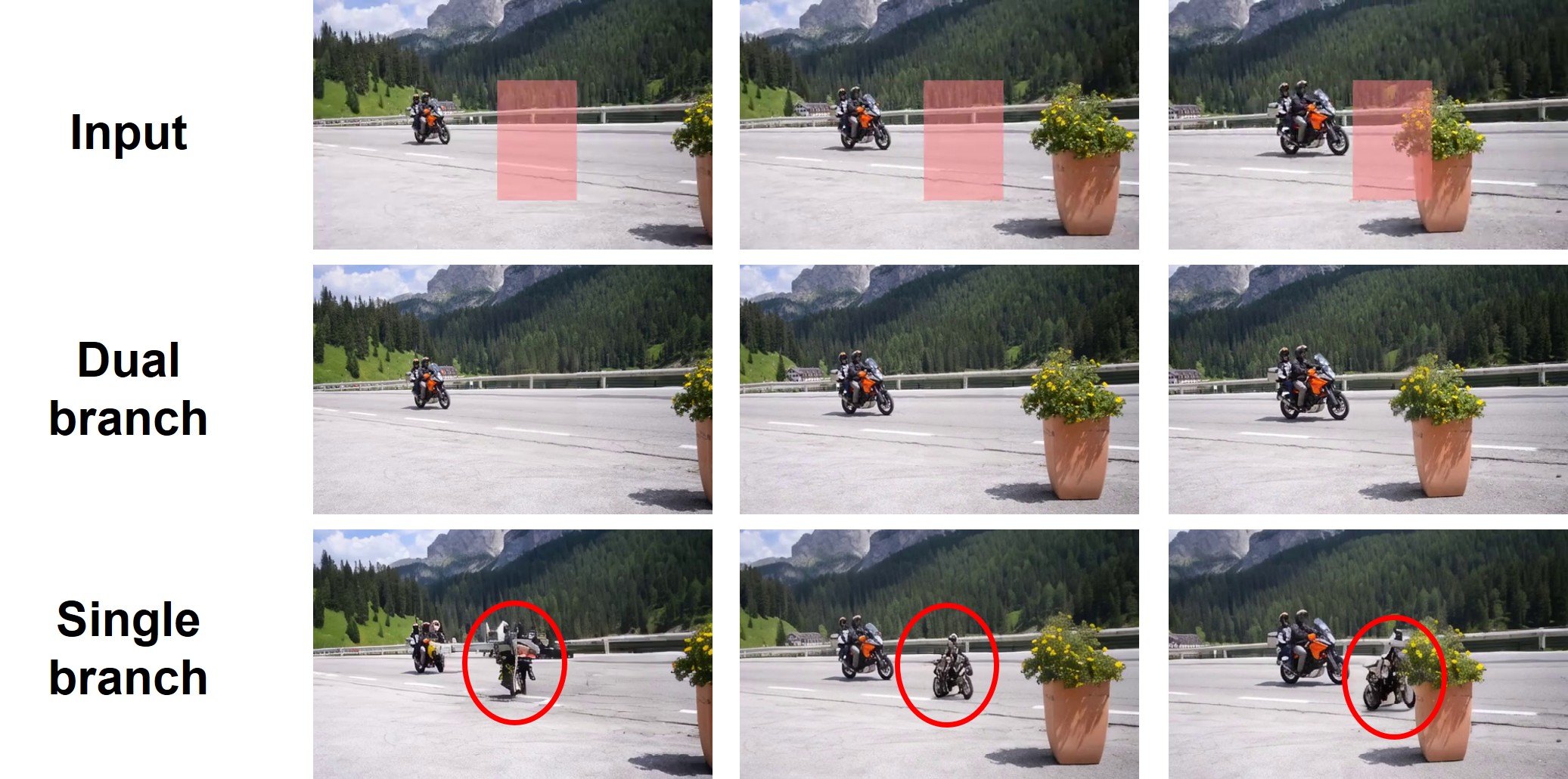}
    \caption{Comparison of dual-branch and single-branch architectures for scene completion tasks. The dual-branch approach effectively separates scene completion from object insertion task. The single-branch model sometimes fails to distinguish between the two tasks, leading to undesired object insertion during scene completion (circled in red).}
    \label{fig.dual}
\end{figure}

As described earlier, our diffusion U-Net utilizes a dual-branch spatial attention mechanism with shared temporal attention to handle both object insertion and scene completion tasks. To assess the necessity of this design, we tested an alternative configuration where both tasks were merged into a single branch, \ie, all spatial and temporal attention layers were shared. The two tasks were trained with equal probability, while all other experimental settings remained consistent.
The results presented in Table~\ref{tab.abl.dual} and Figure~\ref{fig.dual} demonstrate that merging the two tasks into a single branch results in performance degradation for both. We hypothesize that this is due to the conflicting generation objectives of the two tasks, making it more challenging to optimize the network. Furthermore, as shown in Figure~\ref{fig.dual}, when performing scene completion, the single-branch model occasionally inserts objects incorrectly, indicating a failure to distinguish between the two tasks. This observation highlights the necessity of decoupling the tasks with separate branches, allowing each branch to learn task-specific weights effectively.

\begin{table*}[h]
    \centering
    \caption{Comparison between single-branch and dual-branch designs for object insertion and scene completion tasks.}

    \begin{tabular}{l|cccc|cccc}
        \toprule
        & \multicolumn{4}{c|}{Object insertion} & \multicolumn{4}{c}{Scene completion} \\
        \cmidrule{2-9}
        & CLIP-T $\uparrow$ & TempCons $\uparrow$ & mIOU (\%) $\uparrow$ & ImageReward $\uparrow$ 
        & PSNR $\uparrow$ & LPIPS $\downarrow$ & TempCons $\uparrow$ & ImageReward $\uparrow$ \\
        \midrule
        Single-branch & 26.95 & 93.26 & 81.00 & -0.613 & 26.09 & 0.051 & 93.65 & 0.354 \\
        Dual-branch & \textbf{28.78} & \textbf{94.82} & \textbf{85.00} & \textbf{0.106} 
        & \textbf{27.52} & \textbf{0.043} & \textbf{95.02} & \textbf{0.527} \\
        \bottomrule
    \end{tabular}
    \label{tab.abl.dual}
\end{table*}

\subsubsection{Which Layers to Apply Dual-Branch?}

To determine which spatial layers benefit most from the dual-branch design, we experiment with applying the dual-branch mechanism to only self-attention layers, only cross-attention layers, and both self- and cross-attention layers.
We evaluate these three different configurations under the same experimental settings on scene completion task.
The results are presented in Table~\ref{tab.abl.attn}. Our observations indicate that applying the dual-branch design to both self- and cross-attention layers yields the best reconstruction performance, as evidenced by higher PSNR and lower LPIPS values.

\begin{table}[h]
    \centering
    \setlength{\tabcolsep}{0.5mm}
    \caption{Ablation study on applying the dual-branch design to self-attention, cross-attention, or both.}
    \begin{tabular}{cc|cccc|c}
        \toprule
        Self & Cross & PSNR $\uparrow$ & LPIPS $\downarrow$ & TCons $\uparrow$ & ImgReward $\uparrow$ & $\Delta$ Param \\
        \midrule
        \checkmark & $\times$ & 26.89 & 0.044 & 94.13 & 0.405 & 49.6M (+3.5\%) \\
        $\times$ & \checkmark & 26.74 & 0.045 & 93.96 & \textbf{0.454} & 50.3M (+3.6\%) \\
        \checkmark & \checkmark & \textbf{27.44} & \textbf{0.043} & \textbf{94.14} & 0.424 & 99.9M (+7.1\%) \\
        \bottomrule
    \end{tabular}

    \label{tab.abl.attn}
\end{table}

\subsubsection{Long Video Strategy}

To handle long videos of arbitrary length, in addition to our proposed pipeline described in Section~\ref{sec.meth.long}, we implement and evaluate the following strategies:

\begin{itemize}[leftmargin=*]
    \item Direct-gen: Directly inpainting all frames at once.
    \item Multi-gen: Dividing the original long video into multiple overlapped sub-clips, denoising these clips simultaneously at each timestep, and the overlapped frames are averaged. This strategy was proposed by MultiDiffusion~\cite{multidiff} and adopted by AVID~\cite{avid}.
    \item Recur-I2V: Splitting the original video into several non-overlapping clips. The first clip is inpainted first, and subsequent clips are inpainted recurrently in I2V mode, conditioned on the last frame of the previous clip.
    \item Keyframe + in-between inpainting: our default strategy.
\end{itemize}

\begin{figure}
    \centering
    \includegraphics[width=1\linewidth]{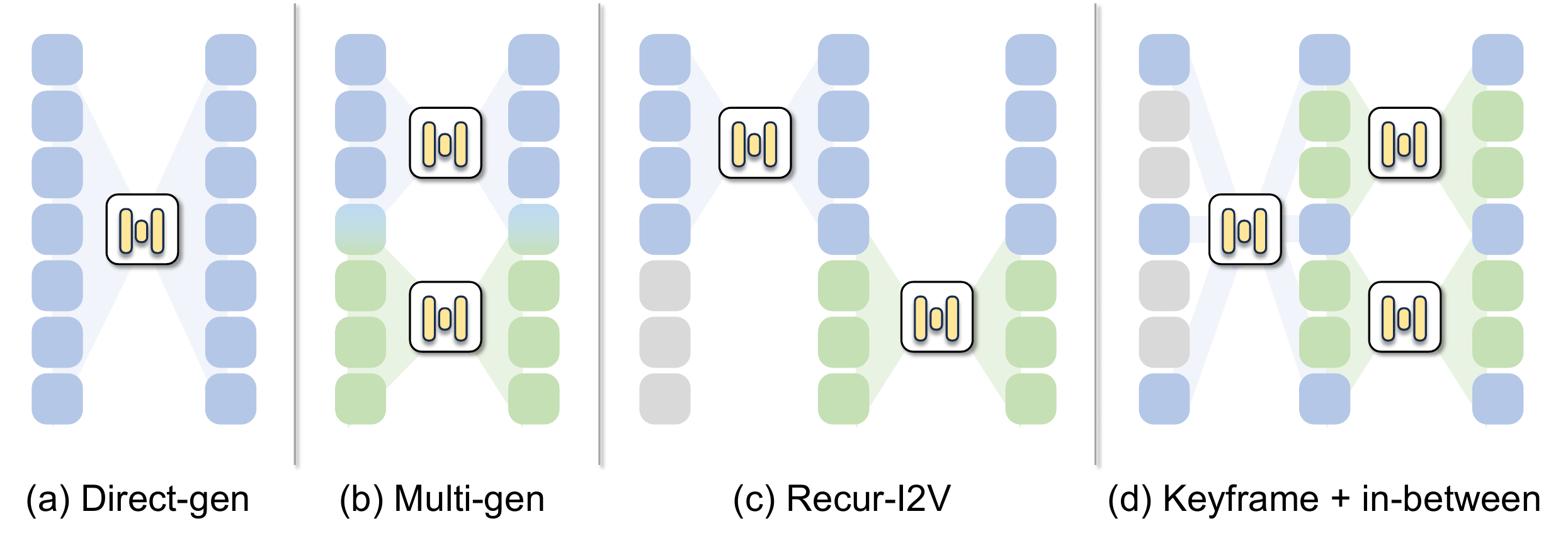}
    \caption{Illustration of different long video generation strategies.}
    \label{fig.long}
\end{figure}

Figure~\ref{fig.long} briefly illustrates the workflow of these strategies.
We evaluate these strategies on the object insertion task using full-length videos in the test set (70 frames on average). In Table~\ref{tab.abl.long}, we primarily report two metrics: in addition to the CLIP-T score, we calculate the average CLIP image similarity between each frame and \textit{the first frame}, denoted as TempCons-F1. This metric specifically measures the consistency of the object across long video sequences. 

In Figure~\ref{fig.long_example}, we show two visual examples from these strategies.
Multi-gen exhibits the worst temporal consistency, as the correlation between frames is easily lost when sub-clips are far apart (evidenced by the varied patterns highlighted with red boxes in Figure~\ref{fig.long_example}). Direct-gen shows lower text fidelity due to a domain gap; the model has not been trained on such long sequences. Additionally, Direct-gen is prone to out-of-memory issues when handling very long videos. Recur-I2V suffers from error accumulation because the information flow is unidirectional, relying solely on the previous frames. In contrast, our default strategy provides bidirectional references for intermediate frames, resulting in the best temporal consistency.

\begin{table}[h]
    \centering
    \caption{Comparison of different long video generation strategies on the object insertion task.}
    \begin{tabular}{l|cc}
        \toprule
        Strategy & CLIP-T $\uparrow$ & TempCons-F1 $\uparrow$ \\
        \midrule
        Direct-gen & 28.44 & 90.18 \\
        Multi-gen & 28.73 & 88.40 \\
        Recur-I2V & 28.70 & 90.15 \\
        Keyframe+in-between (default) & \textbf{28.75} & \textbf{90.59} \\
        \bottomrule
    \end{tabular}

    \label{tab.abl.long}
\end{table}

\subsubsection{Prior Noise Initialization.}

\begin{figure}
    \centering
    \includegraphics[width=0.8\linewidth]{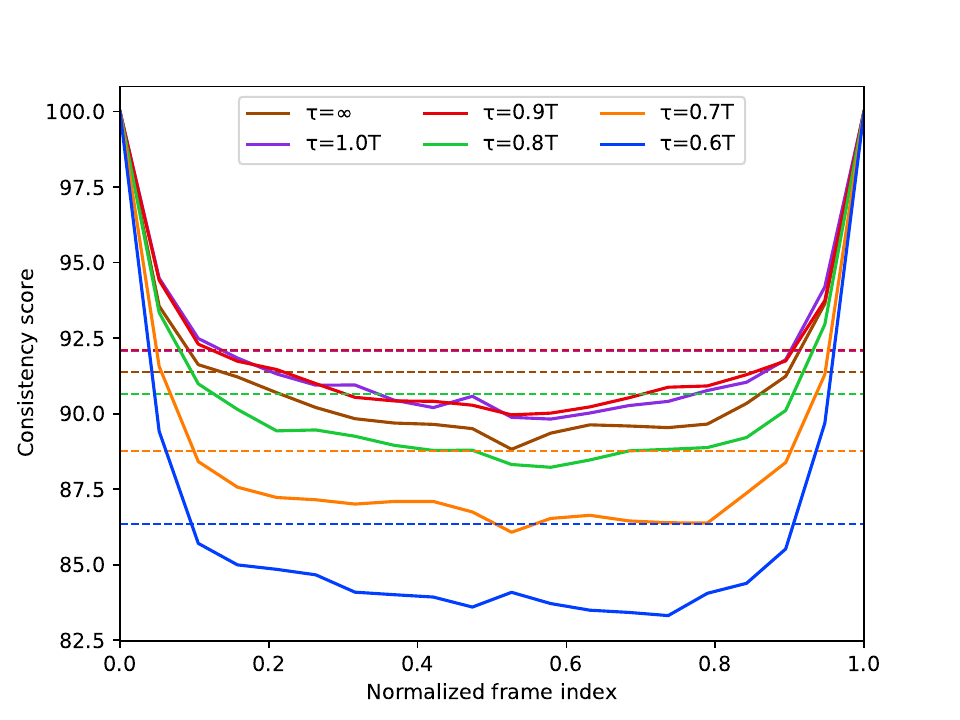}
    \caption{Effect of different noising timestep $\tau$ on frame consistency, dashed lines represent mean values.}
    \label{fig.noise_init}
\end{figure}

To validate the effectiveness of our prior noise initialization (Section~\ref{meth.long.noise_init}), we perform K2V inpainting experiments, where the intermediate frames are inpainted given the first and last frames. Specifically, we adjust the noising timestep $\tau$ and calculate the consistency score between the intermediate frames and the first or last frame, whichever is closer. The results are shown in Figure~\ref{fig.noise_init}, where solid lines represent the consistency scores and dashed lines indicate the mean values.
As observed, setting $\tau \in [0.9T, 1.0T]$ generally yields the highest consistency ($T$ is the number of sampling timesteps). Based on this observation, we select $\tau = 0.9T$ as the default value for our experiments.

\section{Other Applications} \label{sec.app}
Benefiting from our training schemes, \sysname can be extended to additional applications without modifying the model architecture. Here we show several applications including multi-modal inpainting, object editing, object removal, and image object brush.

\subsection{Multi-Modal Inpainting} \label{sec.app.mm}
\begin{figure}
    \centering
    \includegraphics[width=1\linewidth]{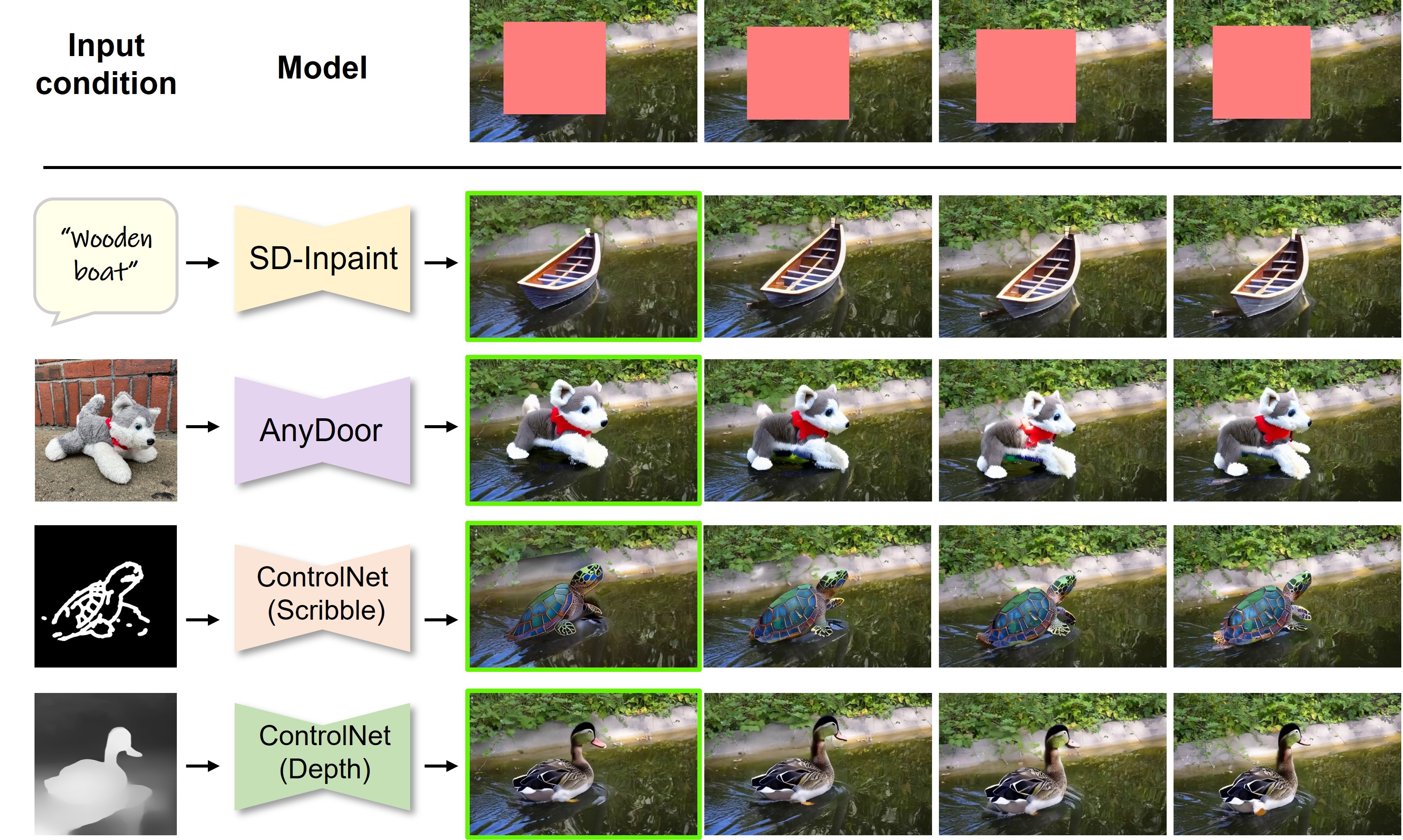}
    \caption{By integrating our method with existing image inpainting model, our framework allows multi-modal guided video inpainting}
    \label{fig.mm}
\end{figure}

Our method stands out from previous approaches by enabling the integration of existing powerful image inpainting tools to perform multi-modal guided video inpainting. We demonstrate such versatility in Figure~\ref{fig.mm}. 
Row 2 showcases text-guided inpainting via SD-inpaint~\cite{ldm}. Row 3 demonstrates exemplar-guided inpainting using AnyDoor~\cite{anydoor}. Rows 4 and 5 utilize ControlNet~\cite{controlnet} for scribble-guided and depth-guided inpainting, respectively. This flexibility empowers users to meet their customized needs, making our framework adaptable for a variety of practical applications.

\subsection{Object Editing} \label{sec.app.object_edit}
\begin{figure*}
    \centering
    \includegraphics[width=0.97\linewidth]{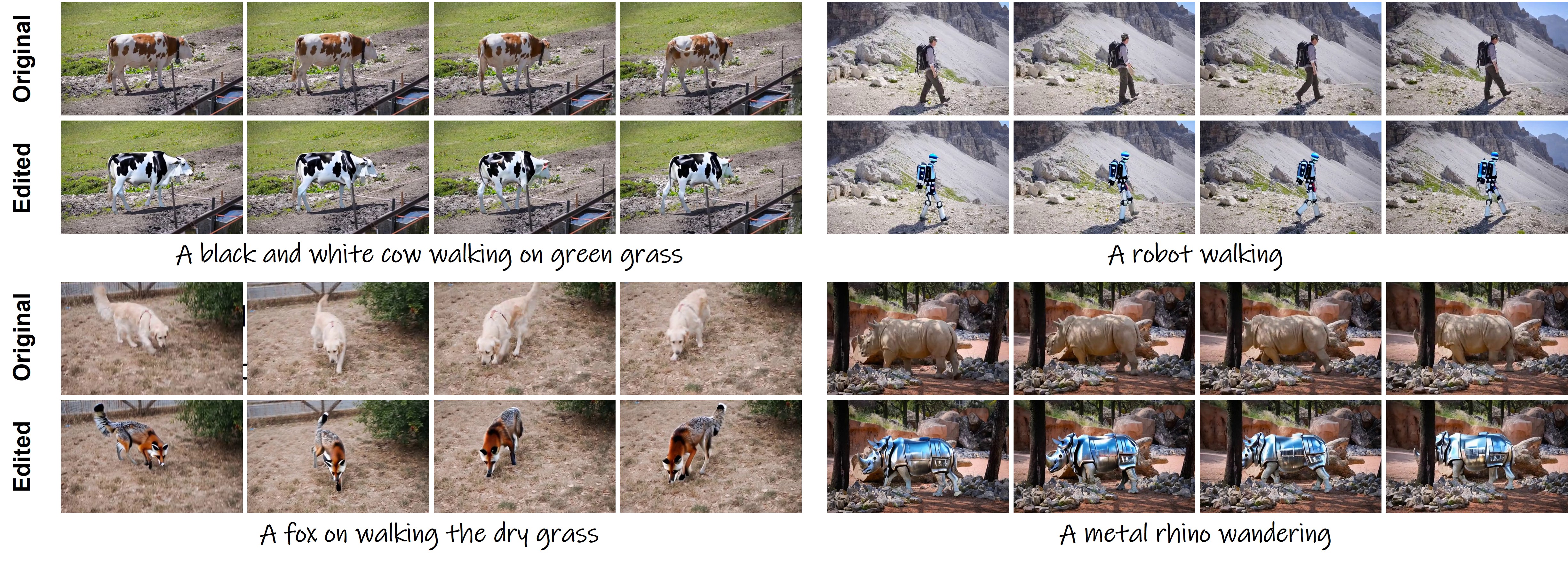}
    \caption{Object editing results. We recommend watching our supplementary video for dynamic results.}
    \label{fig.object_edit}
\end{figure*}

Object editing refers to altering the appearance of an object while preserving its original structure in a source video. Our model demonstrates this capability in a zero-shot manner, inherited directly from the object insertion task. Specifically, instead of using box masks, we can utilize precise object masks during inference and run the insertion branch with an editing prompt. These precise masks provide a strong shape prior, enabling the model to retain the original shape of the object while modifying its appearance.
Figure~\ref{fig.object_edit} illustrates several examples of edited videos, showcasing how our approach effectively maintains the structural integrity of the original object while applying the desired edits.

\subsection{Object Removal} \label{sec.app.object_removal}
\begin{figure*}
    \centering
    \includegraphics[width=0.97\linewidth]{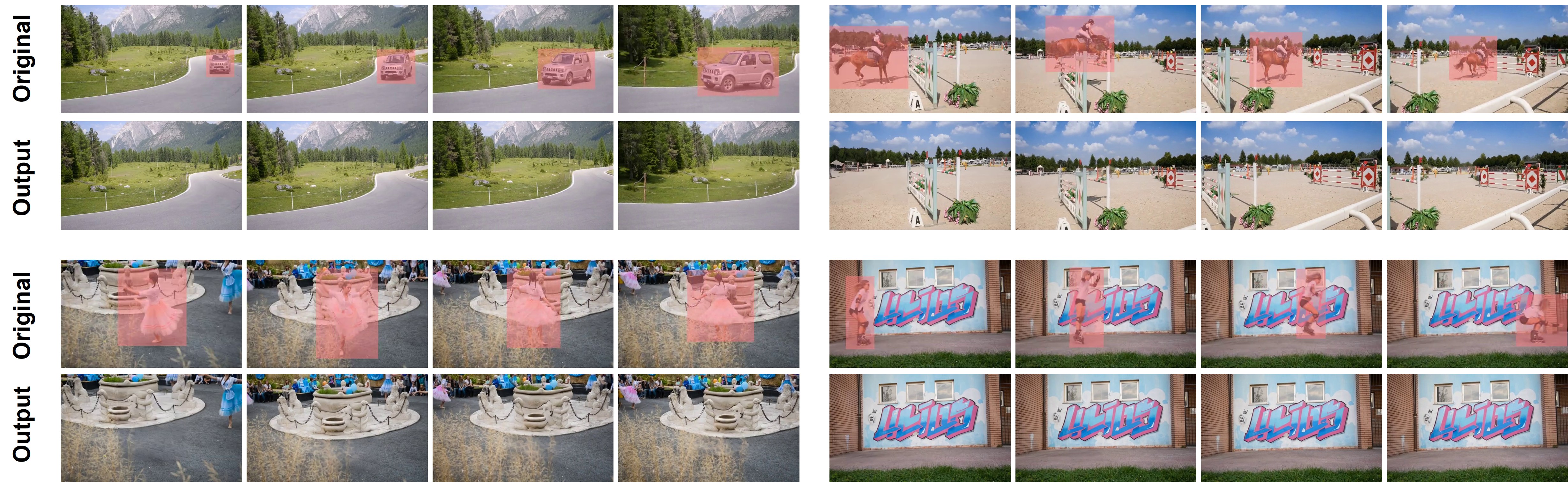}
    \caption{Object removal results. We recommend watching our supplementary video for dynamic results.}
    \label{fig.object_removal}
\end{figure*}

Object removal involves replacing a foreground object with background content, which is a specific case of scene completion. The key difference lies in the requirement for the masks to completely cover the object, ensuring that no part of the foreground object is fed into the model during the inpainting process.
Figure~\ref{fig.object_removal} presents several examples, demonstrating how our method effectively removes the moving foreground object while maintaining the coherence of the background.

\subsection{Image Object Brush} \label{sec.app.object_brush}
\begin{figure*}
    \centering
    \includegraphics[width=0.97\linewidth]{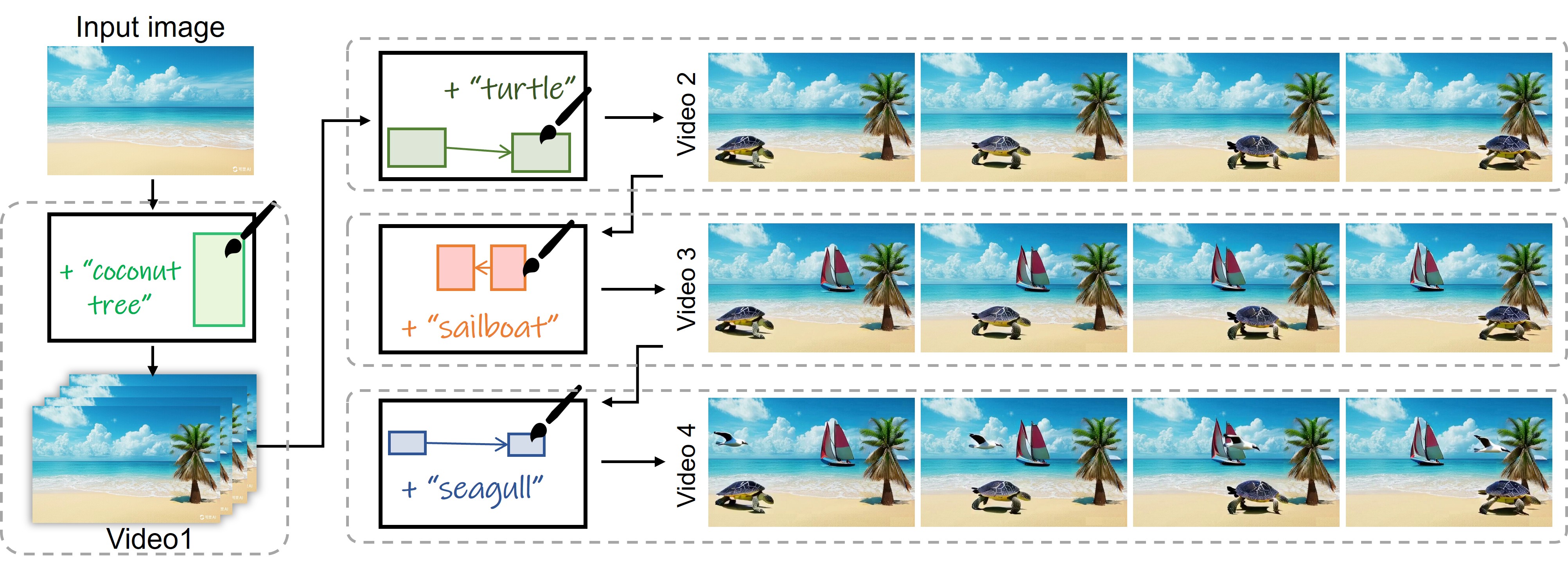}
    \caption{Image object brush: users can draw box trajectories to iteratively add objects into static image to make a dynamic video. We recommend watching our supplementary video for dynamic results.}
    \label{fig.object_brush}
\end{figure*}

Image object brush refers to inserting an animated object into a static image scene by drawing box trajectories. The object can be specified either by a text prompt or a reference image. Similar approaches have been explored in previous studies, such as DragNUWA~\cite{dragnvwa} and Motion-I2V~\cite{motioni2v}, which utilize key point dragging to control object motion. Our framework naturally extends this concept by supporting box dragging, which is more expressive than sparse point-based methods. Box dragging not only controls the motion but also specifies the object’s size, offering enhanced flexibility.
Figure~\ref{fig.object_brush} demonstrates an example where we iteratively add animated objects into a static scene to make a video containing multiple objects.

\section{Limitation and Discussion}
\label{sec.limit}

\begin{figure}
    \centering
    \includegraphics[width=\linewidth]{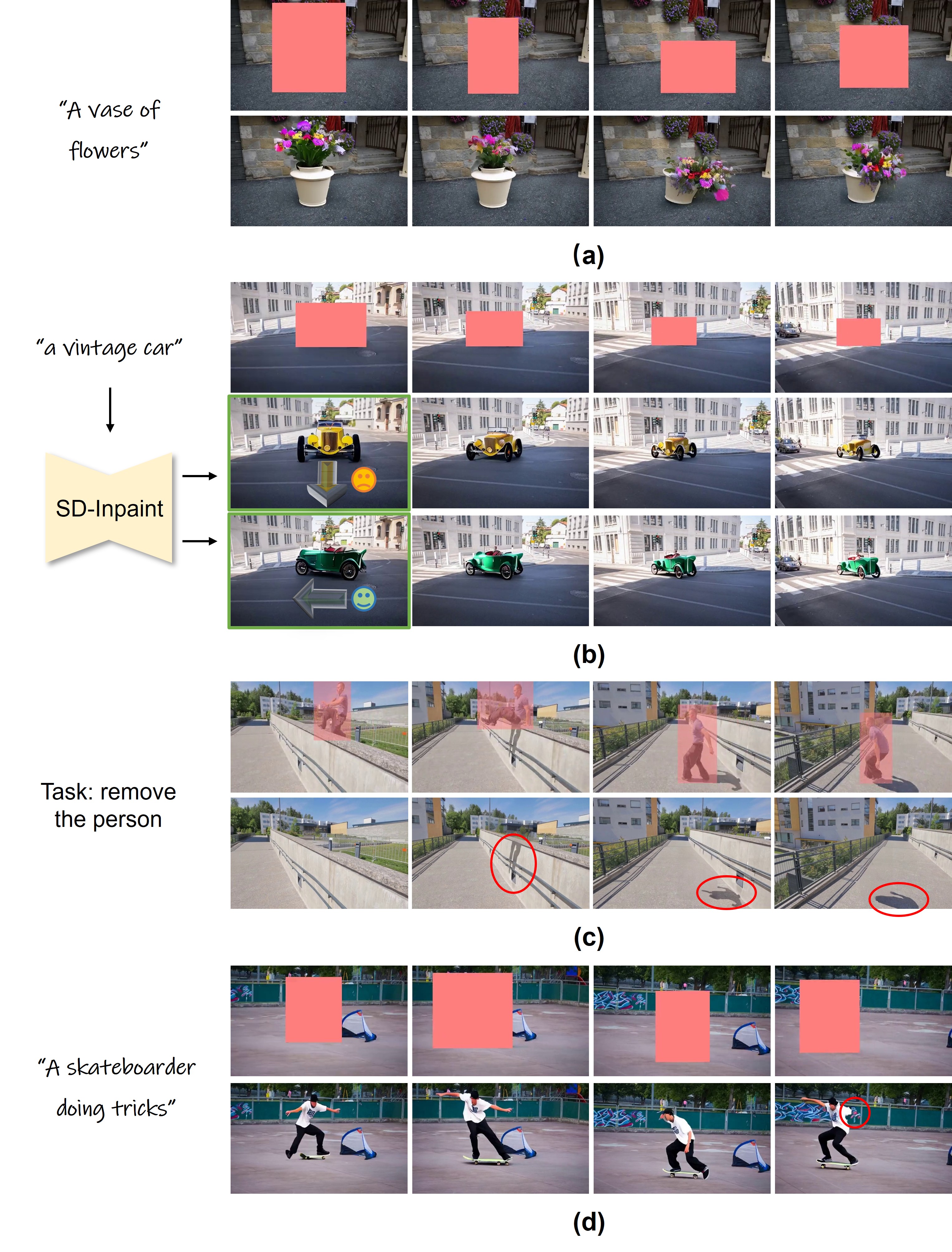}
    \caption{Examples of failure cases. \textbf{(a)} Text-guided object insertion with a moving mask for a static object (vase) leads to unrealistic motion following the mask trajectory. \textbf{(b)} Image-guided object insertion using third-party image inpainting models may occasionally generate a first frame with incorrect object orientation, leading to unnatural motion, such as the lateral sliding of the car shown in the 2nd row. \textbf{(c)} Our object removal results will leave unrealistic residual shadows (highlighted in the red circles) when the mask fails to capture object's shadow regions. \textbf{(d)} Limited synthesis capabilities for complex motions (skateboarding tricks) due to base model constraints.}
    \label{fig.limit}
\end{figure}

Although our approach offers comprehensive solutions for multimodal video inpainting, it is important to acknowledge several limitations.

Firstly, when performing text-guided object insertion, conflicts may arise if a user attempts to insert a stationary object (e.g., a vase of flower) but provides a moving mask as the motion guidance signal, as shown in Figure~\ref{fig.limit}(a). Given these contradictory inputs, our model generates an output where the vase moves with the mask, resulting in an unrealistic results. This issue can be mitigated through appropriate user interaction. Additionally, inserting a stationary object into a dynamic scene requires a mask sequence where the spatial position of the mask changes across frames. Since our design requires users to provide the mask sequence, this can introduce interaction challenges. As this is primarily an engineering issue rather than a core limitation of our framework, we plan to address it in future work by estimating the mask trajectory from user-defined mask in the first frame.

Secondly, when performing image-guided object insertion, we rely on third-party image inpainting models to complete the first frame. However, image inpainting models lack temporal awareness, and the inpainted image may not always be suitable as the first frame for video sequence. For example, in Figure~\ref{fig.limit}(b), the user provides a mask sequence indicating a car moving to the left. However, the inpainting model may generate a first frame where the car is oriented frontally (2nd row) instead of facing left (3rd row), leading to an unrealistic lateral sliding motion. To address this, additional controllable conditions could be incorporated to define the initial pose, or vision-language models could be employed to filter suitable first-frame candidates from a batch of generated results.

Thirdly, in the object removal task, our method relies on object tracking models to determine the mask area. However, these models often fail to include the shadows left behind by the objects, leading to shadow artifacts in the removal results, as indicated by red circle in Figure~\ref{fig.limit}(c). This highlights the need for more advanced tracking models capable of identifying shadows as well.

Finally, our inpainting capabilities are inherently constrained by the capacity of the underlying T2V base model. Complex motions, like "a skateboarder doing tricks" in Figure~\ref{fig.limit}(d), can be challenging for the base model to synthesize, our model inherits such limitation. As a result, the inpainted human body exhibits reduced fidelity in such scenarios. However, it is worth noting that our framework is not tied to a specific model architecture. Future improvements could involve applying our framework to more powerful base models, which we leave as future work.

\section{Conclusion}
\label{sec.conclusion}

In this work, we present \sysname, a multi-task video inpainting framework that addresses the challenges of object insertion, scene completion, and long video inpainting. 
Unlike previous approaches, our method integrates dual-branch spatial attention and shared temporal attention to handle these distinct tasks within a single framework. This design ensures object consistency and dynamic scene reconstruction simultaneously. 
To enhance controllability, we bridge video inpainting with powerful image inpainting tools through the I2V inpainting mode, enabling flexible and multimodal inputs, such as text, exemplars, and edge maps.
For long video inpainting, we propose a two-stage pipeline consisting of keyframe inpainting followed by in-between inpainting, ensuring smooth temporal transitions across extended video lengths.
Experimental results demonstrate that \sysname achieves state-of-the-art performance in both object insertion and scene completion tasks. Additionally, we show its versatility in derived applications, such as object removal, editing, and motion brushing. These contributions highlight the potential of \sysname as an adaptable video inpainting tool for diverse practical scenarios.


\bibliographystyle{ACM-Reference-Format}
\bibliography{_main}

\appendix

\end{document}